\documentclass{article} 
\usepackage{iclr2027_conference,times}

\usepackage{amsmath,amsfonts,bm}

\def\eqref#1{equation~\ref{#1}}

\def\1{\bm{1}}

\DeclareMathAlphabet{\mathsfit}{\encodingdefault}{\sfdefault}{m}{sl}
\SetMathAlphabet{\mathsfit}{bold}{\encodingdefault}{\sfdefault}{bx}{n}

\usepackage{hyperref}
\usepackage{url}
\usepackage{graphicx}
\usepackage{booktabs}

\title{Momentum: Persistent Dynamics for Video World Models}

\author{%
  \parbox{\textwidth}{\raggedright
    \vspace{1em}
    Nedko Savov\textsuperscript{1}\thanks{Corresponding author: \texttt{nedko.savov@insait.ai}}  ,\;
    Stefan Maria Ailuro\textsuperscript{1},\\
    Danda Pani Paudel\textsuperscript{1},\;
    Yilun Du\textsuperscript{2},\;
    Xi Wang\textsuperscript{3,4,5},\;
    Luc Van Gool\textsuperscript{1}
    \\[0.5em]
    {\normalfont
    \textsuperscript{1}INSAIT, Sofia University "St. Kliment Ohridski" \quad
    \textsuperscript{2}Harvard University \quad
    \textsuperscript{3}University of Ulm \quad \\
    \textsuperscript{4}TUM \quad
    \textsuperscript{5}MCML - Munich Center for Machine Learning}
  }
}

\newcommand{\methodname}{Momentum}
\newcommand{\worldupdate}{WorldUpdate}
\newcommand{\worldinteract}{WorldInteract}

\definecolor{mygreen}{RGB}{0,128,0}

\iclrfinalcopy 
\begin{document}

\maketitle
\fancyhead{}

\begin{abstract}
Video world models must capture how a scene evolves beyond the current camera view to generate consistent observations when its content reappears. Existing approaches rely on observation history, static point clouds, or image caches constructed from past frames. These representations are often expensive and, more importantly, cannot update dynamic scene content outside the field of view: generated vehicles and pedestrians may be forgotten after leaving the field of view or reappear in physically inaccurate states. We represent an evolving scene through synchronized videos from reference-camera rigs, each comprising eight cameras that jointly cover the surrounding scene. We connect these rigs in a graph and propagate video generation between neighboring rigs to coordinate scene dynamics across locations and time.
We introduce \methodname{}, a two-stage pipeline: \worldupdate{} jointly generates reference-camera videos as a shared memory of scene dynamics, and \worldinteract{} grounds camera-controlled video generation in this evolving memory, allowing moving content to remain represented outside the target view and reappear consistently.
To support training and evaluation, we develop a CARLA pipeline that captures synchronized videos from reference camera rigs and a moving target camera, together with an evaluation framework measuring consistency within the generated world state and the target observations.
Experiments on generated street scenes show that grounding target videos in scene dynamics generated by \methodname{} improves dynamic scene consistency by $126.2\%$ in unseen scenes over static image conditioning, while also improving visual fidelity. Our models and source code will be made publicly available. 

\end{abstract}


\begin{figure}[t]
  \centering
  \includegraphics[width=\linewidth]{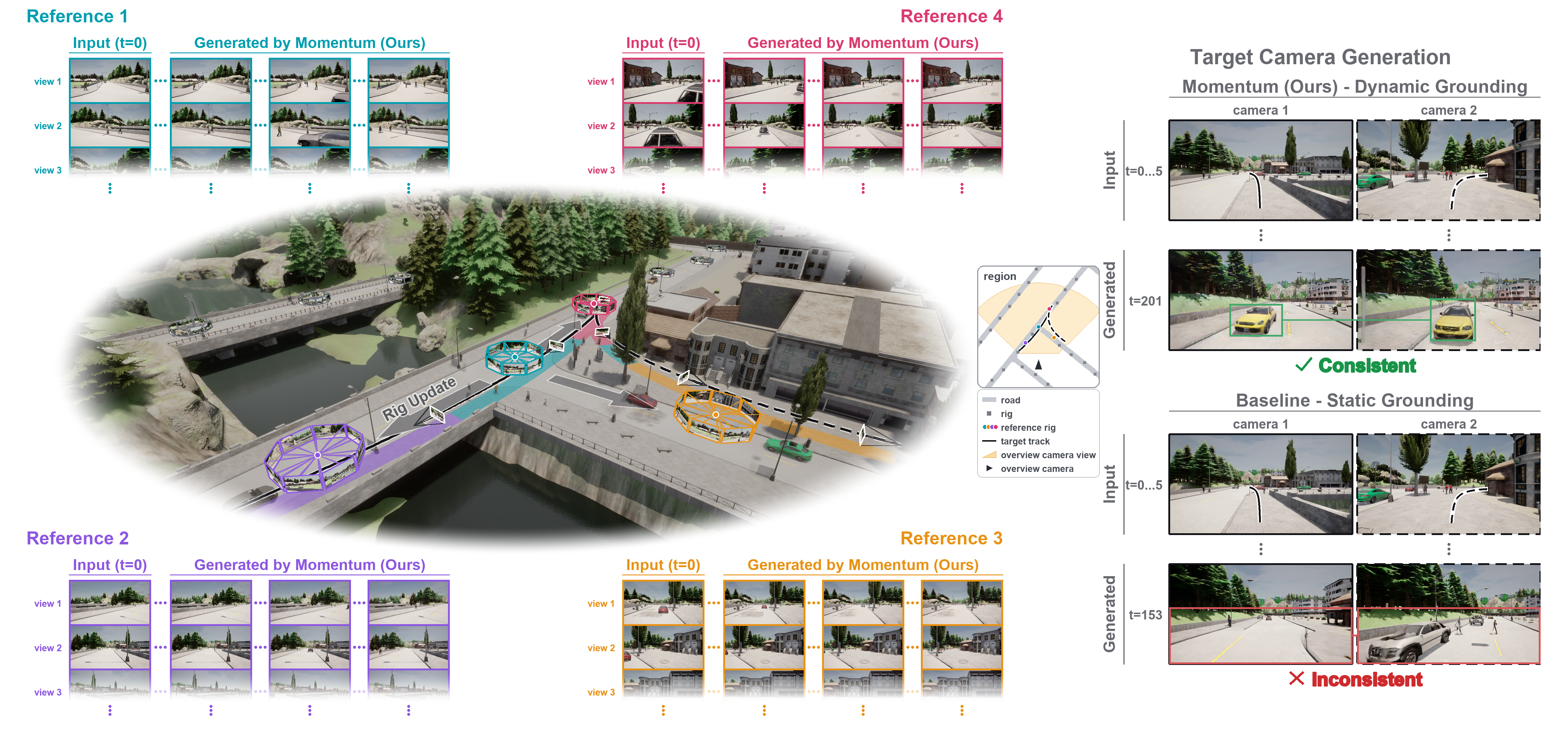}
  \caption{\textbf{Momentum generates temporally and cross-view consistent target
  cameras from sparse camera rig references.} 
  An intersection is observed by four
  8-camera reference rigs (\emph{References 1--4}, colour-coded in the scene
  and shown over time in the corner panels); grey rigs mark further panoramas in the
  region that this example does not condition on. 
  Two moving target cameras
  (black line tracks) traverse the junction, 
  each grounded at every step by its nearest
  reference, indicated by the colour of the band on the road beneath it. Right:
  camera views generated over time for our method and a baseline, from the same $t{=}0$
  ground-truth frame. Inset: plan view of the region with the footprint of the
  overview camera.
  }
  \label{fig:teaser}
\end{figure}

\section{Introduction}

Video world models simulate how visual environments evolve over time and respond to actions, supporting applications in game generation \citep{team2026advancing,yu2025gamefactorly,qian2026matrix}, robotic learning and control \citep{gao2026dreamdojo,ye2026world}, autonomous driving simulation \citep{wayve2026gaia4,basant2026nvidia}, and virtual-world exploration \citep{ball2025genie,team2026advancing,sun2025worldplay}.  In these settings, the generated video reveals only a partial view of an environment whose contents and dynamics extend beyond what is currently visible. As the viewpoint changes, the generated observations should remain consistent with both the scene's spatial structure and the motion of objects within it.
A vehicle approaching an intersection should follow the same course whether the generated video follows it, shows it from another viewpoint, or turns away and later shows it again. This requires grounding generation to a world representation that evolves through time.

Existing memory mechanisms provide part of this grounding through observation history or persistent spatial representations. Conditioning on past frames preserves local continuity, but a finite context limits how much of the environment remains accessible \citep{zhang2026frame,savov2026statespacediffuser,cai2026mixture}. Image caches~\citep{xiao2026worldmem,chen2026learning,yu2025context}, depth maps~\citep{ren2025gen3c,shen2026lyra}, and reconstructed point clouds~\citep{ren2025gen3c,li2025vmem} extend spatial memory by retaining information from previously observed locations. However, stored observations describe the scene at the time they were captured. Without a mechanism for advancing their contents, they cannot provide temporally updated grounding for a moving camera. The challenge is therefore to maintain a representation that specifies both what exists at a location and how its state changes over time.

Recent approaches have begun to model dynamics beyond the current camera view through explicit object-level motion plans or entity-centered simulation within a static background \citep{wang2026worlddirector,duan2026liveworld}. These advances support persistent dynamics for selected entities, while maintaining the evolution of general scene content remains relatively unexplored. We pursue a representation that evolves scene content across a coordinated network of reference cameras, without requiring explicit object trajectories, identity tracking, or a static-dynamic scene decomposition.

Seoul World Model (SWM) offers a useful foundation through image-based spatial grounding~\citep{seo2026seoul}. It conditions camera-controlled video generation on retrieved street-view images, using geometric reprojection and semantic referencing to anchor generated observations to a particular environment. These references ground the scene's spatial structure and appearance, while motion is synthesized within the generated video. We extend this principle to \emph{spatiotemporal grounding}: synchronized videos from spatially distributed reference cameras provide a shared account of both scene appearance and motion. The target camera is thus grounded in the scene at the corresponding location and time, allowing different camera trajectories to observe the same unfolding dynamics.

We introduce \methodname{}, a two-stage pipeline in which \worldupdate{} generates an evolving world state and \worldinteract{} uses it to ground camera-controlled video generation. The method is presented on Fig. \ref{fig:teaser}. The state is organized as a graph of panoramic rigs, each comprising eight reference cameras.
\worldupdate{} jointly generates videos for the eight cameras of each rig and propagates generation through the graph, conditioning each rig on geometrically reprojected video from an already generated neighbor. This couples the evolution of scene content across reference viewpoints. \worldinteract{} then generates video along a target camera trajectory using the resulting world rollout as spatiotemporal grounding. The reference videos therefore serve both as the simulation of scene dynamics and as their persistent representation, enabling motion outside the target camera's view to remain part of the state used to generate subsequent observations.

To develop this pipeline, we build an SWM-based image-grounded model on Cosmos Predict 2.5 and a CARLA data-generation framework that provides training and evaluation data for both stages. The framework captures static reference images and synchronized videos from fixed reference-camera rigs to supervise world-state generation. It also records videos along diverse moving target-camera trajectories, providing supervision for camera-controlled generation grounded in the reference observations. These observations provide supervision for generating a dynamic world state and for rendering camera-controlled video consistent with that state. We develop an evaluation framework for dynamic grounding, combining visual, camera-control, and scene-consistency metrics to assess the evolving world representation and the consistency of target videos with its dynamics.
We evaluate our SWM-based baseline on real-world driving data and assess dynamic grounding on CARLA scenes from seen and unseen towns. Generating reference videos introduces an additional inference stage, trading computation for a shared representation of scene dynamics beyond the current view. This cost can be amortized across multiple target-camera trajectories grounded in the same world rollout, while improving dynamic consistency over static image grounding. With ground-truth reference videos, the Dynamically Grounded WM improves visual fidelity, camera control, and scene consistency over static grounding across single- and multi-rig settings. End-to-end, \methodname{} generates its own reference videos and improves dynamic scene consistency by $130.2\%$ in seen towns and $126.2\%$ in an unseen town compared with the Dynamically Grounded WM conditioned on static images.

Our contributions are threefold:
\begin{itemize}
    \item We introduce a spatiotemporal world representation comprising a graph of reference-camera rigs with synchronized multi-view videos, providing shared spatial and dynamic grounding across camera viewpoints and time.
    \item We propose \methodname{}, a two-stage pipeline that generates scene dynamics through graph-propagated multi-view generation in \worldupdate{} and grounds camera-controlled observations in those dynamics through \worldinteract{}.
    \item We develop a CARLA data-generation pipeline for synchronized reference and target camera observations, an evaluation framework for assessing dynamic scene consistency in both the generated world representation and the resulting target videos, and an image-grounded world model based on SWM.
\end{itemize}

\section{Related Work}
\subsection{Video World Models}Building on learned models of environment dynamics \citep{ha2018world}, generative video world models increasingly support interactive visual environments across domains. Genie \citep{bruce2024genie,ball2025genie}, LingBot-World \citep{team2026advancing}, and HY-WorldPlay \citep{sun2025worldplay} enable controllable exploration. DIAMOND \citep{alonso2024diffusion}, GameFactory \citep{yu2025gamefactorly}, and Matrix-Game 3.5 \citep{qian2026matrix} develop interactive game-like simulation, while Pandora \citep{xiang2024pandora} uses natural-language actions to guide video evolution. Driving models such as GAIA \citep{wayve2026gaia4} and OmniDreams \citep{basant2026nvidia} generate controllable observations for simulation and evaluation. In robotics, DreamDojo \citep{gao2026dreamdojo} learns action-conditioned simulation from human-video priors, DreamZero \citep{ye2026world} jointly predicts future video and robot actions for control, and RoboWorld \citep{jeon2026roboworld} supports robot-policy evaluation through video simulation. Cosmos \citep{agarwal2025cosmos,agarwal2026cosmos} provides pretrained world-model capabilities for physical AI; we build on Cosmos Predict 2.5 \citep{ali2025world}. Our focus is on organizing generated dynamics into a persistent world state that grounds observations across target-camera trajectories.

\subsection{Memory and Dynamics in Video World Models}

Memory mechanisms extend generation beyond a short observation window. StateSpaceDiffuser \citep{savov2026statespacediffuser} uses recurrent state-space modeling, FramePack \citep{zhang2026frame} compresses frame history, and Mixture of Contexts \citep{cai2026mixture} learns sparse attention over informative temporal contexts. WorldMem \citep{xiao2026worldmem}, VRAG \citep{chen2026learning}, Context-as-Memory \citep{yu2025context}, and MemCam \citep{gao2026memcam} retrieve historical observations. Spatial organization further supports consistency through 3D caches in GEN3C \citep{ren2025gen3c}, surfel indexing in VMem \citep{li2025vmem}, and geometric memory in SPMem \citep{wu2026video}, Captain Safari \citep{chou2026captain}, and Lyra 2.0 \citep{shen2026lyra}. Latent Spatial Memory \citep{wang2026latent} stores spatial information in diffusion latent space, while SWM \citep{seo2026seoul} grounds generation in external street-view images. These mechanisms retain past evidence, but memory retention alone does not advance the represented scene through time.

Dynamic geometric methods such as NVFi \citep{li2023nvfi}, FreeGave \citep{li2025freegave}, and TRACE \citep{li2025trace} predict future RGB views by fitting scene dynamics to recorded videos. We instead reuse a dynamics model across scenes and spatial regions without scene-specific fitting. InfiniCube \citep{lu2025infinicube} and WorldSplat \citep{zhu2026worldsplat} generate dynamic environments under supplied object-motion controls; our model generates scene evolution without prescribed object trajectories.

Persistent dynamics beyond the target view remain relatively unexplored. HyDRA \citep{chen2026out} retrieves motion and appearance cues for hidden subjects, while MosaicMem \citep{yu2026mosaicmem} and Mem-World \citep{zheng2026mem} combine spatial memory with dynamic video generation. WorldDirector \citep{wang2026worlddirector} plans object trajectories through an LLM, and LiveWorld \citep{duan2026liveworld} advances active entities through virtual monitors against a static background. \methodname{} instead organizes its evolving state by spatial regions: coordinated reference-camera videos generate and represent scene dynamics, then ground target-camera observations. This formulation supports spatial extension through additional rigs without explicit entity tracking or a static--dynamic decomposition.

\section{Methodology}

\begin{figure}[t]
  \centering
  \includegraphics[width=\linewidth]{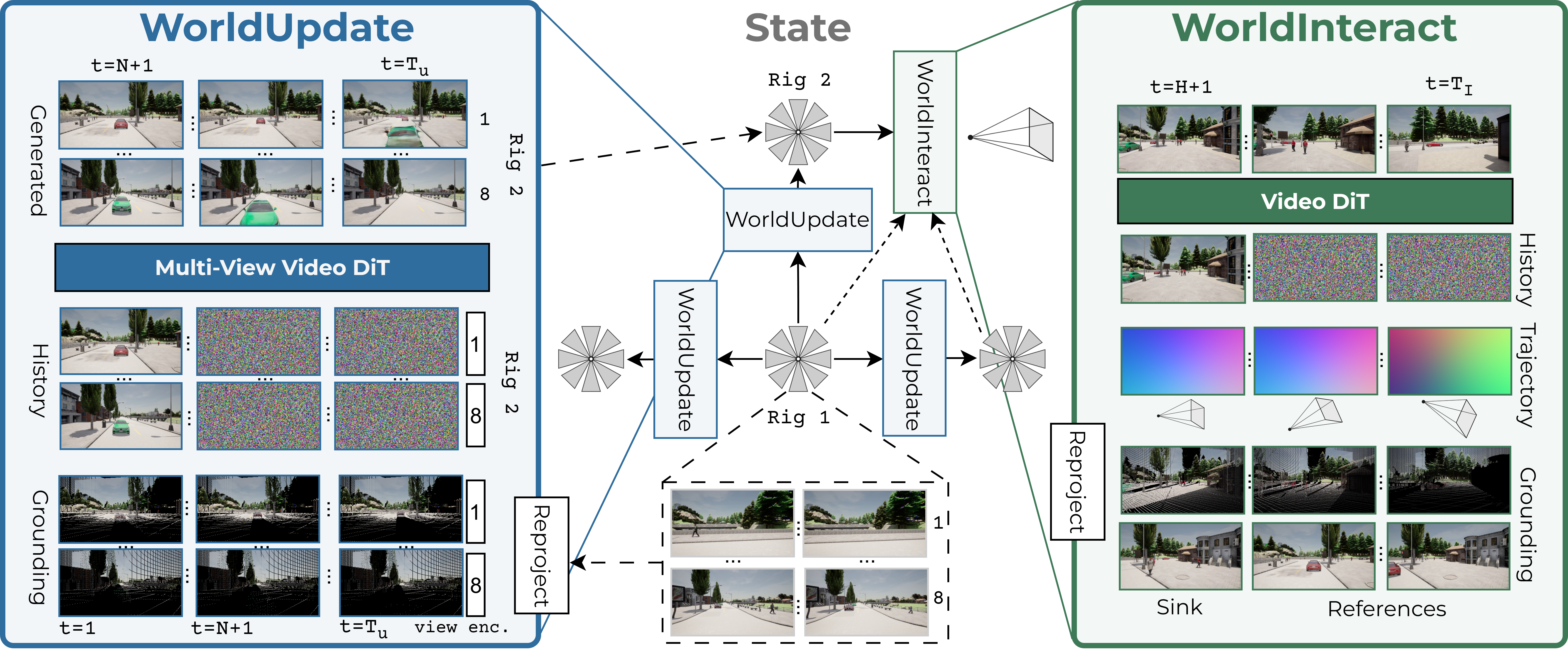}
  \caption{\textbf{Architecture of \methodname{}.}
\worldupdate{} generates synchronized reference-camera videos as an evolving world state, which \worldinteract{} uses to ground target-camera generation. }
  \label{fig:arch}
\end{figure}
\subsection{Problem Statement}
Given a history of \(H\) observations \(I_{1:H}\), the corresponding camera poses \(c_{1:H}\), future camera trajectory \(c_{H+1:H+T}\), and a world grounding \(G\), a grounded camera-controlled video world model \(f(I_{1:H},c_{1:H+T},G)\) predicts the corresponding future observations \(\hat{I}_{H+1:H+T}\), where \(G\) provides additional information about the scene beyond that contained in the observation history, such as spatial context or previously unobserved regions of the environment.

Existing grounded world models such as SWM use \(G\) as a collection of static spatial references, which may originate from different points in time and therefore do not constitute a temporally evolving representation of the dynamic scene. We instead model the grounding as a time-indexed world state \(S_{H:H+T}\), allowing the grounding itself to evolve throughout generation. The prediction can then be written as \(\hat{I}_{H+1:H+T}=h(I_{1:H},c_{H+1:H+T},S_{H:H+T})\), where the state sequence is obtained from an initial grounded state through \(S_{H:H+T}=g(S_H)\).

In our model, \(S_t\) is represented by \(N\) spatially distributed reference-camera rigs, each comprising eight co-located cameras with uniformly spaced orientations. Thus, \(S_t=\{I_t^{n,k}\}_{n=1,\ldots,N;\,k=1,\ldots,8}\). Unlike the static spatial references used in SWM, all reference cameras are temporally synchronized and generate video, so \(S_{H:H+T}\) forms a spatiotemporally evolving representation of the scene.

\subsection{Data Generation Framework}
\label{sec:data}

Training and evaluating our models requires reference observations paired with target-camera trajectories, with synchronized video for dynamic grounding. No large-scale public corpus meets these requirements, and SWM's real-world data is static and proprietary. We therefore develop an open CARLA~\citep{dosovitskiy2017carla} framework for static and dynamic grounding, across six towns with randomized weather, illumination, vehicles, and pedestrians. Further details - in App.~\ref{sec:data_impl}.

\paragraph{Static grounding data.}
We pair target videos with panoramic reference rigs covering the same maps. Each rig contains eight outward-facing cameras spaced $45^\circ$ apart in yaw, positioned $2$\,m above the road. Rigs follow the road graph at $20$\,m intervals with small random positional offsets. Captures are separated by $2.4$\,s, equivalent to $30$\,km/h travel, allowing dynamic content to evolve between neighboring captures as in street-view acquisition. Target videos span $36$\,s at $10$\,fps along vehicle (autopilot), pedestrian (AI-controlled walker), or flyover (freely moving elevated camera) trajectories. References and targets are captured in independent sessions with independently sampled conditions, sharing static structure while differing in transient content.

\paragraph{Dynamic grounding data.}
To supervise evolving reference views, we capture reference and target videos in a shared simulation rollout. A moving target camera traverses a randomly selected intersection covered by one panoramic rig (\emph{single-rig}) or up to four spatially distributed rigs (\emph{multi-rig}). Each session spans $300$ frames at $10$\,fps, with synchronized observations depicting the same scene evolution. To reduce storage, rigs record new observations only when the target camera is within $25$\,m; otherwise, their latest frames are repeated to retain temporal alignment. Active intervals provide eight synchronized streams per rig for world-state supervision, while target trajectories revisit shared content from different viewpoints and headings to supervise grounded camera-controlled generation. 

We collect $20$ sessions per setting per town for training and two evaluation sets per setting of $30$ sessions each: Test~1 uses unseen intersections in training towns, and Test~2 uses a town excluded from training to assess generalization.

\subsection{Static Grounding}
\label{sec:static_wm}

Our \textbf{\textit{Statically Grounded World Model}} follows the
SWM architecture~\citep{seo2026seoul}. As it is currently not publicly available, we build an open implementation of it. Details appear in App.~\ref{sec:swm_impl}.

\paragraph{Backbone and rollout.}
The backbone is a Diffusion Transformer operating in the latent space of a causal 3D VAE, initialized from Cosmos Predict 2.5-2B~\citep{ali2025world}. Generation proceeds autoregressively in chunks, denoising chunk $i$ conditioned on a text prompt $P^{(i)}$ (we use a constant preset prompt), target-camera trajectory $C^{(i)}$, and $H$ history latents from chunk $i-1$. Per-frame Pl\"ucker ray maps, encoded by a shallow convolutional encoder and added to latent tokens, provide camera poses for history, target, reference, and sink latents. We train under teacher forcing on the CARLA data of Sec.~\ref{sec:data}.

\paragraph{Retrieval-augmented grounding.}
Each chunk retrieves up to $K$ reference images near the target trajectory, with camera poses and metric depth supplied by CARLA during training and estimated with Depth Anything V3~\citep{da3} at inference. Two pathways incorporate these references. \emph{Geometric referencing} unprojects the nearest references for each target frame into a point cloud and forward-splats it into the target viewpoint; the resulting warped video is encoded and concatenated channel-wise with noisy target latents for dense spatial conditioning. \emph{Semantic referencing} encodes each reference into a single latent, appended to the token sequence at a RoPE temporal position separated from the generation window by $G_{\mathrm{ref}}$. Target latents attend to all $K$ references to recover appearance information lost through depth errors or occlusion.

A \emph{virtual lookahead sink} adds a retrieved latent at RoPE position $\Delta_{\mathrm{VL}}$ beyond the chunk as a clean anchor, refreshed each generation step. References and targets come from different scene realizations, encouraging reliance on persistent structure rather than transient content. Sec.~\ref{sec:baseline_validation} compares this baseline with SWM's reported scores.
\subsection{\methodname}
\label{sec:momentum}

\methodname{} is a two-stage pipeline that generates an evolving world state and uses it to ground camera-controlled video. The architecture of our method is shown on Fig. \ref{fig:arch}. The state $S_t$ comprises the reference-camera rig views, advanced through time rather than held fixed. \worldupdate{} implements the state transition $S_{H:H+T}=g(S_H)$, generating synchronized videos across rigs to represent scene dynamics independently of the target-camera trajectory. \worldinteract{} implements the observation model $h$, generating target-camera video conditioned on the observation history, camera trajectory, and corresponding world state. The reference videos thus serve both as the simulation of scene dynamics and as their persistent representation, grounding observations when moving content leaves and later re-enters the target view.

Dynamics are represented without explicit entity detection, identity tracking, or re-identification, so for a fixed rig configuration the generation workload does not grow with the number of moving entities. The two stages operate independently: we store completed world rollouts in the dataset format of Sec.~\ref{sec:data}, allowing \worldinteract{} to consume generated and recorded reference videos through the same interface. A single world rollout can ground multiple target-camera trajectories, be inspected independently, and be generated at a different resolution or on a different machine from the target video.

\subsubsection{\worldupdate}
\label{sec:worldupdate}

\worldupdate{} is the component of \methodname{} responsible for advancing the world state through synchronized reference-camera video generation. It requires consistency both across cameras within a rig and across spatially distributed rigs observing shared content. To meet these requirements, we introduce the \emph{\textbf{Multi-Rig Dynamically Grounded World Model}} (Multi-Rig DGWM) as an architecture for this component, combining joint multi-view generation with geometric conditioning between rigs.

\paragraph{Multi-view generation.}
Multi-Rig DGWM builds on the pretrained multi-view video diffusion transformer of Cosmos-Predict 2.5 \citep{ali2025world}, trained on autonomous driving data. It extends its seven-view configuration to the eight cameras of our rigs by adding an extra learnable view embedding. View-stacked attention exchanges information across cameras during joint generation, supporting cross-view consistency. We incorporate the warped-video channels of geometric referencing and the Pl\"ucker camera residual from the grounded model. Semantic reference tokens, the virtual lookahead sink, and their associated attention masks are omitted: each fixed camera's generated history anchors its viewpoint and appearance, while reprojected neighbor videos provide spatially aligned conditioning on scene evolution.

\paragraph{Multi-rig consistency.}
Independent rig generation can produce conflicting depictions of shared content, including objects duplicating or disappearing between viewpoints. We therefore condition each rig on a neighbor's synchronized video, unprojecting its views with depth and reprojecting them into the receiving rig using the known relative poses. During training, conditioning comes from the ground-truth video of the nearest other rig, allowing training samples to be processed independently. At inference, neighboring videos must themselves be generated, requiring a schedule that makes each conditioning source available before its dependent rig.

\paragraph{Graph propagation.}
We construct a graph connecting rigs close within a range to provide useful reprojection conditioning. Generation proceeds in chunks of one model window, overlapping by the conditioning frames to maintain temporal continuity. Within each chunk, we traverse the graph breadth-first from a source rig, which generates from its own history without neighbor conditioning. Each subsequent rig conditions on the freshly generated video of its nearest already-generated neighbor(s), while rigs at equal depth are generated together. This propagates scene evolution across the graph over a shared temporal interval. 


\subsubsection{\worldinteract}
\label{sec:worldinteract}

\worldinteract{} is the component of \methodname{} that generates target-camera video grounded in the evolving world state. It must follow the specified camera trajectory while preserving the scene appearance and motion represented by the reference videos. For this component, we introduce the \emph{\textbf{Dynamically Grounded World Model}} (Dynamically Grounded WM), retaining the architecture of the Statically Grounded World Model (Sec.~\ref{sec:static_wm}) and fine-tuning it on synchronized dynamic grounding data (Sec.~\ref{sec:data}).

For each chunk, the model retrieves $K$ nearby rig views as semantic references and reprojects the nearest reference view into each target frame. Unlike static grounding, geometric conditioning uses the reference video frame at the corresponding target time, conveying the scene's evolving appearance and motion. Semantic references are instead sampled at the chunk's start, providing a common temporal anchor across the retrieved viewpoints. 

\begin{table}[t]
\caption{\textbf{Baseline   on Ann-Arbor-City-Bench (MARS).} SWM scores are those reported by the authors; our statically grounded model is trained on synthetic data alone. Masked PSNR and LPIPS are computed on static regions only, following the SWM protocol. Best value per metric in bold.}
\label{tab:baseline_validation}
\begin{center}
\small
\setlength{\tabcolsep}{3.2pt}
\begin{tabular}{lcccc}
\multicolumn{1}{c}{\bf Method}  &\multicolumn{1}{c}{\bf mPSNR $\uparrow$}  &\multicolumn{1}{c}{\bf mLPIPS $\downarrow$}  &\multicolumn{1}{c}{\bf RotErr $\downarrow$}  &\multicolumn{1}{c}{\bf TransErr $\downarrow$}
\\ \midrule
SWM~\cite{seo2026seoul} (reported)                  & 15.18          & 0.481          & 0.055          & \textbf{0.154} \\
Statically Grounded World Model (ours)    & \textbf{16.30} & \textbf{0.426} & \textbf{0.025} & 0.238          \\
\end{tabular}
\end{center}
\caption{\textbf{\worldinteract{} consistency across scene configurations.} Models use ground-truth reference observations from scenes containing one or four rigs. Image conditioning uses fixed initial observations; video conditioning uses synchronized frames. Best scores for each scene configuration are \textbf{bold}; best scores across configurations within each test set are \underline{underlined}.}
\label{tab:world_update_consistency_rigsinscene}
\resizebox{\linewidth}{!}{%
\setlength{\tabcolsep}{3.2pt}
\begin{tabular}{ll*{12}{c}}
\multicolumn{1}{c}{\bf Method} & \multicolumn{1}{c}{\bf Cond.}
& \multicolumn{2}{c}{\bf PSNR $\uparrow$}
& \multicolumn{2}{c}{\bf LPIPS $\downarrow$}
& \multicolumn{2}{c}{\bf RotErr $\downarrow$}
& \multicolumn{2}{c}{\bf TransErr $\downarrow$}
& \multicolumn{2}{c}{\bf SCS-static $\uparrow$}
& \multicolumn{2}{c}{\bf SCS-dyn $\uparrow$} \\
\multicolumn{2}{r}{rigs in the scene:}
& 1 rig & 4 rigs
& 1 rig & 4 rigs
& 1 rig & 4 rigs
& 1 rig & 4 rigs
& 1 rig & 4 rigs
& 1 rig & 4 rigs 
\\ \midrule
\multicolumn{14}{l}{\textit{test 1: seen towns}} \\
Statically Grounded WM & image
& 21.01 & 19.89 
& 0.436 & 0.426
& 0.031 & 0.038
& 0.181 & 0.160
& 0.512 & 0.526
& 0.056 & 0.078 \\
Statically Grounded WM & video
& 20.97 & 20.35
& 0.436 & 0.414
& 0.029 & 0.030
& 0.171 & 0.156
& 0.502 & 0.562
& 0.083 & 0.127 \\
Dynamically Grounded WM (ours) & video
& \underline{\textbf{23.55}} & \textbf{23.16}
& \textbf{0.352} & \underline{\textbf{0.335}}
& \underline{\textbf{0.021}} & \textbf{0.022}
& \underline{\textbf{0.054}} & \textbf{0.056}
& \textbf{0.630} & \underline{\textbf{0.652}}
& \textbf{0.202} & \underline{\textbf{0.265}}
\\ \midrule
\multicolumn{14}{l}{\textit{test 2: unseen town}} \\
Statically Grounded WM & image
& 17.94 & 20.27
& 0.508 & 0.449
& 0.026 & 0.026
& 0.176 & 0.148
& 0.426 & 0.573
& 0.057 & 0.081 \\
Statically Grounded WM & video
& 17.93 & 19.23
& 0.507 & 0.456
& 0.026 & 0.034
& 0.196 & 0.167
& 0.425 & 0.570
& 0.059 & 0.112 \\
Dynamically Grounded WM (ours) & video
& \textbf{20.36} & \underline{\textbf{21.94}}
& \textbf{0.425} & \underline{\textbf{0.371}}
& \underline{\textbf{0.023}} & \underline{\textbf{0.023}}
& \underline{\textbf{0.047}} & \textbf{0.053}
& \textbf{0.583} & \underline{\textbf{0.667}}
& \textbf{0.147} & \underline{\textbf{0.230}} \\

\end{tabular}%
}
\end{table}
\begin{table}[t]
\centering
\caption{\textbf{\worldinteract{} consistency across reprojection source counts.} Within four-rig scenes, geometric conditioning combines ground-truth reference observations from one or three source rigs. Best scores for each source count are \textbf{bold}; best scores across source counts within each test set are \underline{underlined}.}
\label{tab:world_update_consistency_rigswarp}
\resizebox{\linewidth}{!}{%
\setlength{\tabcolsep}{3.2pt}
\begin{tabular}{ll*{12}{c}}
\multicolumn{1}{c}{\bf Method} & \multicolumn{1}{c}{\bf Cond.}
& \multicolumn{2}{c}{\bf PSNR $\uparrow$}
& \multicolumn{2}{c}{\bf LPIPS $\downarrow$}
& \multicolumn{2}{c}{\bf RotErr $\downarrow$}
& \multicolumn{2}{c}{\bf TransErr $\downarrow$}
& \multicolumn{2}{c}{\bf SCS-static $\uparrow$}
& \multicolumn{2}{c}{\bf SCS-dyn $\uparrow$} \\
\multicolumn{2}{r}{warp sources:}
& 1 src & 3 src
& 1 src & 3 src
& 1 src & 3 src
& 1 src & 3 src
& 1 src & 3 src
& 1 src & 3 src 
\\ \midrule
\multicolumn{14}{l}{\textit{test 1: seen towns}} \\
Statically Grounded WM & image
& 19.89 & 19.55
& 0.426 & 0.429
& 0.038 & 0.034
& 0.160 & 0.173
& 0.526 & 0.503
& 0.078 & 0.074 \\
Statically Grounded WM & video
& 20.35 & 20.04
& 0.414 & 0.419
& 0.030 & 0.029
& 0.156 & 0.172
& 0.562 & 0.540
& 0.127 & 0.107 \\
Dynamically Grounded WM (ours) & video
& \underline{\textbf{23.16}} & \textbf{22.89}
& \underline{\textbf{0.335}} & \textbf{0.337}
& \textbf{0.022} & \underline{\textbf{0.020}}
& \textbf{0.056} & \underline{\textbf{0.053}}
& \underline{\textbf{0.652}} & \textbf{0.647}
& \underline{\textbf{0.265}} & \textbf{0.230}
\\ \midrule
\multicolumn{14}{l}{\textit{test 2: unseen town}} \\
Statically Grounded WM & image
& 20.27 & 20.33
& 0.449 & 0.449
& 0.026 & 0.026
& 0.148 & 0.146
& 0.573 & 0.563
& 0.081 & 0.079 \\
Statically Grounded WM & video
& 19.23 & 20.42
& 0.456 & 0.430
& 0.034 & 0.027
& 0.167 & 0.143
& 0.570 & 0.589
& 0.112 & 0.131 \\
Dynamically Grounded WM (ours) & video
& \textbf{21.94} & \underline{\textbf{23.05}}
& \textbf{0.371} & \underline{\textbf{0.351}}
& \textbf{0.023} & \underline{\textbf{0.019}}
& \textbf{0.053} & \underline{\textbf{0.043}}
& \textbf{0.667} & \underline{\textbf{0.680}}
& \textbf{0.230} & \underline{\textbf{0.260}} \\

\end{tabular}%
}
\end{table}
\begin{figure}[t]
    \centering
    \includegraphics[width=0.7\linewidth]{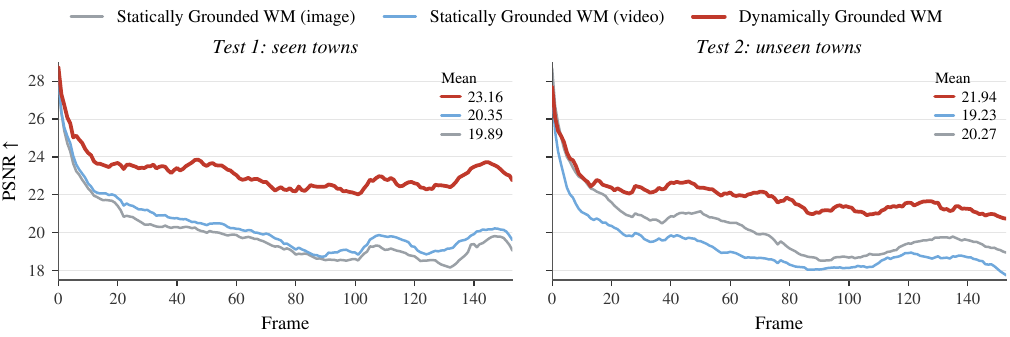}
    \caption{\textbf{PSNR per frame over the generation horizon of \worldinteract{}.}}
    \label{fig:egopsnr}
    \vskip -10pt
\end{figure}
\begin{table}[t]
\caption{\textbf{End-to-end evaluation.} \methodname{} rolls out the static references in the world state with \worldupdate{} and renders the ego view with \worldinteract{}; Statically and Dynamically Grounded models are directly conditioned on the available static image references.}
\label{tab:end_to_end}
\begin{center}
\small
\setlength{\tabcolsep}{3.2pt}
\begin{tabular}{lcccccc}
\multicolumn{1}{c}{\bf Method}  &\multicolumn{1}{c}{\bf PSNR $\uparrow$}  &\multicolumn{1}{c}{\bf LPIPS $\downarrow$}  &\multicolumn{1}{c}{\bf RotErr $\downarrow$}  &\multicolumn{1}{c}{\bf TransErr $\downarrow$}  &\multicolumn{1}{c}{\bf SCS-static $\uparrow$}  &\multicolumn{1}{c}{\bf SCS-dyn $\uparrow$}
\\ \midrule
\multicolumn{7}{l}{\textit{test 1: seen towns}} \\
Statically Grounded WM                     & 19.89 & 0.426 & 0.038 & 0.160 & 0.526 & 0.078 \\
Dynamically Grounded WM (image)       & 22.02 & 0.358 & \textbf{0.026} & \textbf{0.060} & 0.623 & 0.086 \\ 
\methodname{}                           & \textbf{24.03} & \textbf{0.341} & \textbf{0.026} & 0.080 & \textbf{0.658} & \textbf{0.198} \\
\midrule
\multicolumn{7}{l}{\textit{test 2: unseen town}} \\
Statically Grounded WM                     & 20.27 & 0.449 & 0.026 & 0.148 & 0.573 & 0.081 \\
Dynamically Grounded WM (image)       & 22.44 & 0.380 & 0.021 & 0.048 & 0.658 & 0.084 \\ 
\methodname{}                             & \textbf{22.98} & \textbf{0.358} & \textbf{0.018} & \textbf{0.040} & \textbf{0.698} & \textbf{0.190} \\
\end{tabular}
\end{center}
\end{table}

\section{Experiments}

\paragraph{Experimental Setup.}
We train Statically Grounded WM, Dynamically Grounded WM and Multi-Rig DGWM on the CARLA training set (Sec. \ref{sec:data}) on 6 towns, according to the training schedule outlined in App. \ref{sec:swm_impl} and App. \ref{sec:momentum_impl}. We evaluate them in the roles of components of Momentum. We evaluate on Test 1 and Test 2, as collected in Sec. \ref{sec:data}. Unless otherwise stated, we use the 4-rig version of each set and choose 2 views of 1 closest rig for reprojection. We evaluate visual fidelity - PSNR, LPIPS,  camera trajectory
accuracy - rotation and translation error and scene consistency - SCS-static (\cite{bagchi2026walkpaintingsegocentricworld}), which we adapt to dynamic objects in SCS-dyn. See App.~\ref{app:metrics}.

\paragraph{Baseline Validation.}
\label{sec:baseline_validation}
We evaluate our Statically Grounded World Model on the publicly available Ann Arbor driving data from MARS~\citep{Li_2024_CVPR}, following SWM's evaluation protocol, using masked versions of PSNR and LPIPS (see App. \ref{app:metrics}). This comparison assesses whether our reimplementation provides a competitive baseline despite training exclusively on synthetic data, whereas SWM additionally uses video interpolated from proprietary real-world panoramas. As shown in Tab.~\ref{tab:baseline_validation}, our model achieves higher mPSNR, lower mLPIPS, and lower rotation error than the reported SWM results, although translation error remains higher. These results establish a competitive image-grounded baseline for evaluating the benefits of dynamic grounding.

\paragraph{\worldinteract{} Consistency Evaluation.}
We evaluate model choices for \worldinteract{} using ground-truth reference observations, isolating target-video generation from errors in \worldupdate{}. We compare the Statically Grounded World Model conditioned on either fixed reference images from $t=0$ or synchronized reference video against the Dynamically Grounded WM conditioned on video. Image conditioning reprojects the initial observations into each target viewpoint, while video conditioning uses time-aligned observations. Across scenes containing one or four reference-camera rigs, the Dynamically Grounded WM achieves the best visual fidelity, camera-trajectory accuracy, and static and dynamic scene consistency in both seen and unseen towns (Tab.~\ref{tab:world_update_consistency_rigsinscene}). Supplying video conditioning to the statically trained model alone does not yield comparable gains, demonstrating the importance of training with synchronized dynamic grounding.

\paragraph{\worldinteract{} Reprojection Sources.}
We next vary the number of rigs supplying conditioning within four-rig scenes, using either one or three source rigs with the same model and conditioning choices. From each source, we select the two camera views closest in orientation to the target camera and combine their reprojections into a single conditioning image per target frame. The Dynamically Grounded WM outperforms both baselines with either source count (Tab.~\ref{tab:world_update_consistency_rigswarp}). In seen towns, one source provides better visual and semantic consistency, while three slightly improve camera-trajectory accuracy. In the unseen town, three sources improve all metrics, despite training with a single source rig, showing that the model can exploit additional reference coverage at inference.

\paragraph{\worldinteract{} Performance Over Generation Horizon.}
We evaluate how visual fidelity changes throughout generation using per-frame PSNR on both test sets. As shown in Fig.~\ref{fig:egopsnr}, the Dynamically Grounded WM maintains visual fidelity more consistently over time than models using static grounding.

\paragraph{End-to-End Evaluation.}
We evaluate \methodname{} from static reference images, without ground-truth reference videos. The baselines directly condition target-video generation on these images, while \methodname{} uses \worldupdate{} to generate an evolving world state through graph propagation and \worldinteract{} to ground target observations in that state. As shown in Tab.~\ref{tab:end_to_end}, \methodname{} improves visual fidelity and both static and dynamic scene consistency over both baselines in seen and unseen towns. It also achieves the best camera-trajectory accuracy in the unseen town, while matching rotation accuracy but increasing translation error relative to the image-conditioned Dynamically Grounded WM in seen towns. Most notably, SCS-dyn improves over this baseline by $130.2\%$ and $126.2\%$ in seen and unseen towns, respectively, demonstrating the benefit of generating scene dynamics beyond those represented by static references.

\begin{figure}[t]
    \centering
    \includegraphics[width=1\linewidth]{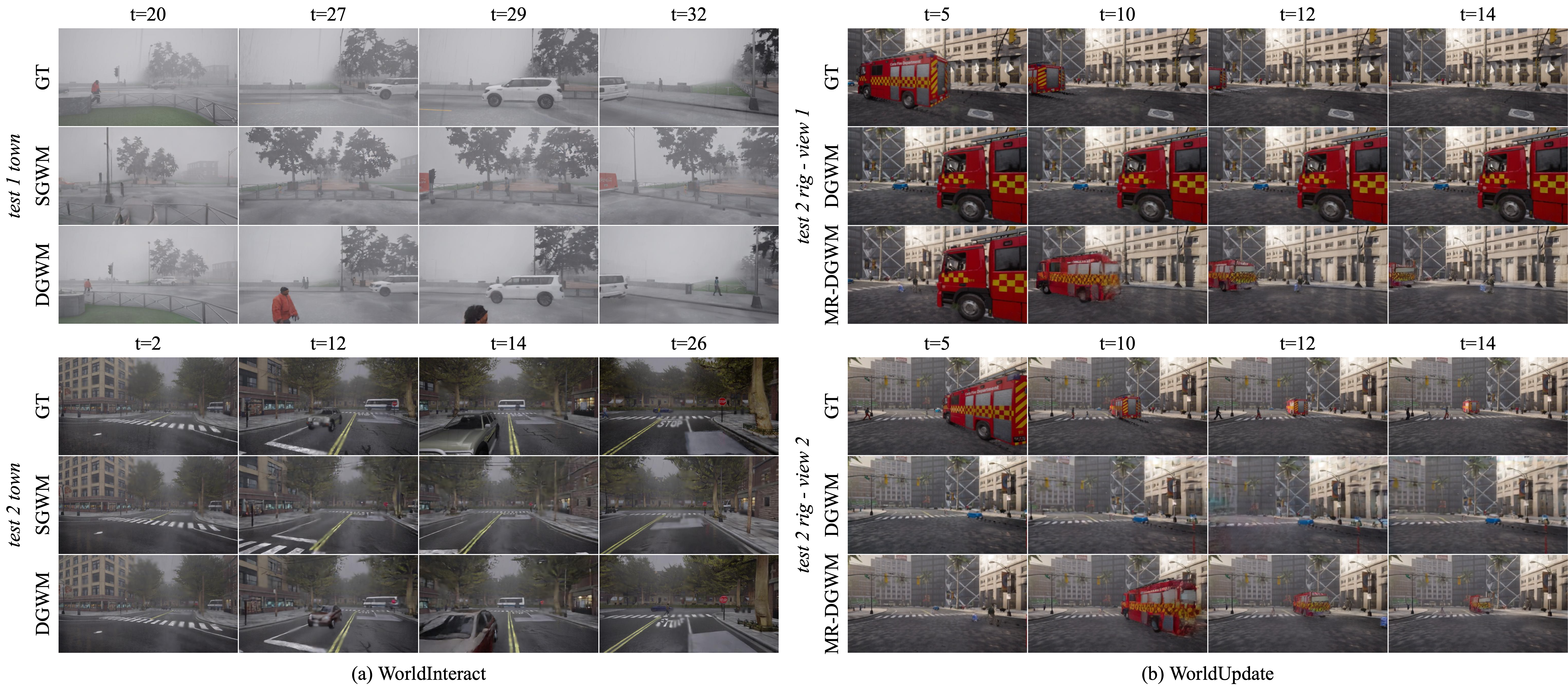}
    \caption{\textbf{Qualitative examples.} a) WorldInteract, SGWM versus DGWM in the scenes of both testsets; b) WorldUpdate, DGWM vs Multi-Rig DGWM in two views of the same rig in the test 2. }
    \label{fig:qual}
\end{figure}

\paragraph{Qualitative Results.}
We qualitatively show examples of \worldupdate{} and \worldinteract{}, comparing static and dynamic grounding, single-view and multi-rig models. In Fig.~\ref{fig:qual}a, it is apparent that the dynamically grounded model preserves moving objects (cars, pedestrians) significantly better across the inference window. In Fig.~\ref{fig:qual}b, the multi-rig model renders objects consistently across views and predicts their movement, while the single-view model is inconsistent and tends to predict no movement (both models are seeded with single frame at t=0).

\section{Conclusion}

We introduced \methodname{}, a two-stage pipeline that grounds camera-controlled video generation in an evolving world state. Synchronized reference-camera videos represent scene appearance and dynamics beyond the target view, without explicit entity tracking or a static--dynamic decomposition. \worldupdate{} advances this state through graph-propagated generation, while \worldinteract{} uses it to ground target-camera observations. We developed the Dynamically Grounded WM and Multi-Rig DGWM for these roles, together with a CARLA data-generation framework, an SWM-based Statically Grounded World Model, and an evaluation framework. Experiments in seen and unseen towns demonstrate improved visual fidelity, controllability and scene consistency through dynamic grounding. Compared with the Dynamically Grounded WM conditioned on static image references, \methodname{} improves dynamic scene consistency (SCS-dyn) by $130.2\%$ in seen towns and $126.2\%$ in the unseen town. These results open a path toward spatially extensible video world models in which multiple viewpoints and interactions share a consistently evolving scene.

\section{AI use statement}

In this work, we used generative AI tools for providing feedback on research  methodology or experiments and implementing methods. We have not used generative AI tools for designing research methodology, generating synthetic datasets,  helping develop theoretical models or conceptual frameworks, supporting qualitative and thematic data analysis, interpreting results, proposing or refining hypotheses. We have not used an AI tool in any form to formulate the ideas and design our system. Formulating mathematical claims, providing critical ingredients for proving mathematical claims, assisting in the writing of proofs, assisting with translation, cleaning and reformatting dataset are not applicable to this work.
Additionally, we used generative AI tools for creating or modifying scientific figures or images,  creating software code, drafting parts of the research paper and improving its readability.
We have reviewed all AI-assisted work, including code and text. We take responsibility for the final content of this work,
including text, claims or artifacts produced with the aid of generative AI.

\bibliography{iclr2027_conference}
\bibliographystyle{iclr2027_conference}

\appendix

\section{Appendix}
\subsection{Data Generation Implementation Details}
\label{sec:data_impl}

All observations are rendered using $1280\times720$ pinhole cameras with a $90^\circ$ field of view. Each capture stores RGB video together with per-frame camera extrinsics, intrinsics, and GNSS measurements; depth is stored losslessly. Data collection is configured declaratively, supports resuming interrupted sessions, and can be distributed across multiple CARLA servers.

\paragraph{Static grounding.}
Panoramic rigs are placed at road waypoints with a random positional offset of up to $1.5$\,m. Each rig contains eight co-located cameras with yaw offsets spaced by $45^\circ$ and records both RGB and depth. Target vehicle trajectories use an autopilot vehicle travelling at a randomized fraction below the local speed limit. Pedestrian trajectories use an eye-height camera at $1.65$\,m mounted to an AI walker with randomly sampled navigation goals. Flyover trajectories follow collision-checked random walks at $2$--$8$\,m altitude, with camera pitch sampled around $-20^\circ$ with $\pm15^\circ$ variation and additional yaw perturbations. We render $20$ sessions per town and trajectory type.

\paragraph{Dynamic grounding.}
Each session is centred on a randomly selected intersection with at least three approach directions, bounding-box half-extent of at least $5$\,m, and sufficient nearby vehicle spawn points and pedestrian navigation area within $60$\,m. In the multi-rig setting, up to four rigs are placed on road waypoints within $40$\,m of the intersection and at least $12$\,m apart, with one rig located at the junction centre. The single-rig setting retains only the central rig.

We spawn up to $10$ vehicles and $25$ pedestrians around the selected region. The target camera is placed at an eye height of $1.8$--$2.2$\,m and follows a smoothed, collision-checked trajectory visiting the rigs in randomized order. The path remains within $30$\,m of the nearest reference rig in the multi-rig setting and $20$\,m in the single-rig setting, promoting repeated observation of shared geometry from different headings, including U-turns.

Each session consists of a single synchronous CARLA rollout of $300$ frames at $10$\,fps. Every reference rig records eight RGB and eight depth streams aligned with the target-camera timeline. To reduce rendering cost, reference rigs are active only while the target camera is within $25$\,m; outside this range, the most recent frame is repeated to preserve temporal alignment. Test sessions use disjoint random seeds, and Test~1 additionally excludes intersections used during training.

\subsection{Statically Grounded World Model Implementation Details}
\label{sec:swm_impl}

\paragraph{Architecture.}
The backbone is the $2$B Cosmos-Predict2.5 DiT: $28$ blocks of width $2048$ with $16$ heads, spatial patch size $2$, AdaLN-LoRA of rank $256$, and 3D RoPE. The VAE compresses by $4\times$ temporally and $8\times$ spatially into $16$ latent channels. Three modifications are made. First, the input embedder is widened from $16$ to $32$ latent channels so that the warped video of the geometric pathway can be concatenated to the noisy target; with the condition and padding masks, the network sees $34$ channels. Second, separate patch embedders, weight-initialised from the base embedder, are added for reference and sink latents. Third, RoPE temporal indices are decoupled from position in the token sequence, allowing references and the lookahead sink to be placed outside the generation window, and attention masks are added for reference and sink padding. All remaining blocks are inherited unchanged.

\paragraph{Chunk geometry.}
Each chunk covers $77$ pixel frames, which the causal VAE encodes to $L=20$ latents, preceded by $H=5$ history latents, for a $25$-latent generation window. We use $K=5$ references, a reference gap of $G_{\mathrm{ref}}=50$ latent positions, an inter-reference spacing of $1$, and a lookahead offset of $\Delta_{\mathrm{VL}}=5$ latent frames. Clean conditioning latents (history, references, and sink) are assigned timestep $0$ under the base model's per-frame AdaLN.

\paragraph{Training.}
We fine-tune from the public Cosmos-Predict2.5-2B checkpoint for $10$K iterations at $704\times1280$ resolution with AdamW ($\mathrm{lr}=4.8\times10^{-5}$, weight decay $0.001$, $\beta=(0.9,0.999)$), $100$ warm-up steps, and gradient clipping at $0.1$, under the base rectified-flow recipe (logit-normal timestep sampling, shift $5$). Training is sharded with FSDP across $8$ NVIDIA H100 GPUs at a per-GPU batch size of $1$. To avoid over-reliance on any single conditioning signal, we perturb history latents with Gaussian noise of standard deviation $0.1$ with probability $0.5$, and drop the reference and warped-video pathways independently with probability $0.2$ each.

\paragraph{Inference.}
Chunks are rolled out sequentially with a rectified-flow ODE sampler using $35$ steps and classifier-free guidance of $3.0$, with only the history latents carried forward; the camera trajectory, retrieved references, and sink are re-derived at every chunk.

\subsection{\methodname{} Implementation Details}
\label{sec:momentum_impl}

\paragraph{\worldupdate{}: Multi-Rig DGWM.}
The rig model is initialised from the released $2$B multi-view Cosmos-Predict2.5 checkpoint, with an eighth view-embedding row appended to its seven-slot table and the warped-video latent appended to the input embedder, giving $41$ input channels ($16$ noisy, $1$ condition mask, $1$ padding mask, $7$ view, $16$ warped). Every pretrained column keeps its index, so the conversion only adds zero-initialised weights. All eight views of a rig are generated jointly at $192\times352$ per view with $8$ latent frames per view, corresponding to $29$ pixel frames per chunk. Generation is conditioned on a single history latent ($h=1$), with the initial chunk conditioned on one ground-truth frame per view of each rig.

We fine-tune for $10$K iterations with AdamW ($\mathrm{lr}=3\times10^{-5}$, weight decay $10^{-3}$, $\beta=(0.9,0.999)$), $100$ warm-up steps, and gradient clipping at $0.1$, sharded with FSDP. The warped latent is dropped with probability $0.2$ per sample, which both teaches the model to generate without a reprojection source and provides the branch without reprojection conditioning for classifier-free guidance. This rate is chosen to approximate how often a rig is the graph source at inference.


\paragraph{\worldinteract{}: Dynamically Grounded World Model.}
The target-camera model is fine-tuned from the statically grounded checkpoint described in App.~\ref{sec:swm_impl} on the dynamic grounding sessions, retaining that model's chunk geometry, optimiser settings, and conditioning dropout. It generates target-camera video at $704\times1280$ resolution in chunks of $77$ pixel frames ($L=20$ latents), conditioned on $H=5$ history latents. At inference, it uses the same $35$-step rectified-flow ODE sampler and classifier-free guidance of $3.0$ as the statically grounded model.

References are bound to the first frame of the chunk's window. The sink is read $\Delta_{\mathrm{VL}}=5$ latent frames past the target end, clamped to the length of the available state. During training, a sample's geometric pathway is disabled unless at least $95\%$ of its target frames fall within a window in which the source rig was actively rendering.

\paragraph{Rig connectivity and source selection.}
Reference rigs actively render within a range of $r=25$\,m. This range is recorded with each session and also determines the edges of the rig graph used by \worldupdate{}. For target-camera generation, \worldinteract{} reprojects the nearest in-range rig by default. Rigs outside the active rendering range hold their last frame and are excluded as reprojection sources.

\section{Metrics}
\label{app:metrics}


\subsection{Camera Trajectory Accuracy}
\paragraph{RotErr \& TransErr}
These measure how well the camera motion of the generated video follows the target trajectory, using the protocol of~\cite{seo2026seoul} (App.~B.1). We estimate per-frame camera-to-world poses $(\mathbf{R}_t, \mathbf{t}_t)$ of \emph{both} the generated and the ground-truth video with Depth Anything V3~\cite{da3} at a processing resolution of $504$\,px. Estimating the reference trajectory with the same model, rather than using the simulator's exact poses, reduces the estimator's systematic bias and places both trajectories in the same camera convention. As in~\cite{seo2026seoul}, videos are processed in non-overlapping chunks of $32$ frames, and each chunk is scored in its own coordinate frame. Within a chunk $c$ with first frame $t_0$, poses are expressed relative to that anchor,
\begin{equation}
  \mathbf{R}^{\mathrm{rel}}_t = \mathbf{R}_{t_0}^{\top}\mathbf{R}_t, \qquad
  \mathbf{t}^{\mathrm{rel}}_t = \mathbf{R}_{t_0}^{\top}\bigl(\mathbf{t}_t - \mathbf{t}_{t_0}\bigr),
\end{equation}
and, since monocular pose estimation is only defined up to scale, each translation trajectory is normalised by its own maximum norm within the chunk, $\bar{\mathbf{t}}_t = \mathbf{t}^{\mathrm{rel}}_t / \max_{t'\in c}\lVert\mathbf{t}^{\mathrm{rel}}_{t'}\rVert_2$. With $\mathcal{F}$ the set of all non-anchor frames over all chunks of a clip, the rotation error is the mean geodesic distance on $\mathrm{SO}(3)$ and the translation error the mean Euclidean distance between normalised translations:
\begin{equation}
  \mathrm{RotErr} = \frac{1}{|\mathcal{F}|}\sum_{t\in\mathcal{F}}
  \arccos\!\left(\frac{\operatorname{tr}\!\bigl(\hat{\mathbf{R}}^{\mathrm{rel}\,\top}_t \mathbf{R}^{\mathrm{rel}}_t\bigr) - 1}{2}\right), 
\end{equation}
\begin{equation}
  \mathrm{TransErr} = \frac{1}{|\mathcal{F}|}\sum_{t\in\mathcal{F}}
  \bigl\lVert \hat{\bar{\mathbf{t}}}_t - \bar{\mathbf{t}}_t \bigr\rVert_2 .
\end{equation}
The anchor frames, whose error is zero by construction, are excluded. RotErr is reported in radians and TransErr is unitless (it is relative to the chunk's extent); lower is better for both.

\subsection{Consistency Metrics}
\paragraph{SCS static \& dynamic}
The Structural Consistency Score~\cite{bagchi2026walkpaintingsegocentricworld} measures whether the generated video keeps the \emph{layout} of the scene, independent of appearance, as the mean mask IoU between corresponding regions of the generated and ground-truth videos:
\begin{equation}
  \mathrm{SCS} = \frac{1}{NT}\sum_{j=1}^{N}\sum_{t=1}^{T}
  \frac{\bigl|\hat{M}^{t,j}\cap M^{t,j}\bigr|}{\bigl|\hat{M}^{t,j}\cup M^{t,j}\bigr|}.
\end{equation}
Masks are generated automatically with SAM\,3.1~\cite{sam3} tracking over a fixed vocabulary of semantic categories, and $M^{t,j}$ is the union of all instances of category $j$ tracked in frame $t$. The static and dynamic categories are:
\begin{itemize}
  \item \textbf{SCS-static}: \{\textit{building}, \textit{road},
        \textit{sidewalk}, \textit{vegetation}, \textit{sky}\}, the persistent scene
        structure, which the model must preserve and move consistently with the camera;
  \item \textbf{SCS-dynamic}: \{\textit{vehicle}, \textit{pedestrian},
        \textit{person}, \textit{bicycle}, \textit{animal}\}, moving entities, which additionally
        test whether the model keeps them present and in the right place.
\end{itemize}
Frames in which a category is empty in \emph{both} videos (IoU $0/0$) are excluded. IoUs are pooled over all valid (clip, category, frame) triples. To bound the cost of SAM\,3.1, both videos are temporally subsampled to 2 fps.


\subsection{Video Fidelity Metrics}
\paragraph{mPSNR \& mLPIPS}
Following~\cite{seo2026seoul}, we measure reconstruction fidelity only on the \emph{static} part of the scene. We reuse SAM\,3.1  to extract dynamic masks, as above, for the generated and the ground-truth video, giving $\hat{D}_t, D_t \in \{0,1\}^{H\times W}$. A pixel counts as static only if it is static in both videos:
\begin{equation}
  S_t = \neg\bigl(\hat{D}_t \lor D_t\bigr).
\end{equation}

mPSNR is the PSNR over all static pixels and colour channels of a clip (peak value $1$):
\begin{equation}
  \mathrm{mPSNR} = -10\log_{10}
  \frac{\sum_{t}\sum_{p} S_t(p)\,\lVert \hat{I}_t(p) - I_t(p)\rVert_2^2}
       {3\sum_{t}\sum_{p} S_t(p)} .
\end{equation}

LPIPS~\cite{lpips} is a distance in feature space and cannot be restricted to an arbitrary pixel set, thus for mLPIPS we fill the dynamic regions of the generated frame with the ground truth, $\tilde{I}_t = S_t \odot \hat{I}_t + (1 - S_t)\odot I_t$, and compute
\begin{equation}
  \mathrm{mLPIPS} = \frac{1}{|\mathcal{T}_S|}\sum_{t\in\mathcal{T}_S}
  \mathrm{LPIPS}_{\text{VGG}}\bigl(\tilde{I}_t, I_t\bigr),
  \qquad \mathcal{T}_S = \{t : \textstyle\sum_p S_t(p) > 0\}.
\end{equation}

\subsection{Measures of Change}
Staticness measures \emph{how much of the scene moves} in a video. The exo rig cameras are fixed, so a change in where an object appears in the image can only come from the object moving. We reuse the SAM\,3.1 tracks of the dynamic categories. Unlike SCS, masks are compared \emph{across time within one video} rather than between the generated and ground-truth videos. For a clip $v$, the staticness is the mean IoU between consecutive frames:
\begin{equation}
  \mathrm{St}(v) = \frac{1}{|\mathcal{P}_v|}\sum_{(t,j)\in\mathcal{P}_v}
  \frac{\bigl|M^{t,j}\cap M^{t+1,j}\bigr|}{\bigl|M^{t,j}\cup M^{t+1,j}\bigr|},
  \qquad \mathcal{P}_v = \bigl\{(t,j) : \bigl|M^{t,j}\cup M^{t+1,j}\bigr| > 0\bigr\}.
\end{equation}
If no agents are present or moving, $\mathrm{St}=1$. The score falls as a larger share of the agents' area moves, appears or disappears between frames. As in SCS, both videos are temporally subsampled to 2\,fps, so $t$ and $t{+}1$ are $0.5$\,s apart. Pairs in which a category is empty in both frames are excluded.

We compute $\mathrm{St}$ for the generated video ($\widehat{\mathrm{St}}$) and for the ground truth ($\mathrm{St}$), and average each over clips. SAM\,3.1 masks jitter slightly from frame to frame, so even a perfectly frozen video scores somewhat below $1$; the persistent baseline value is therefore the reference, not $1$. A model can match the average staticness while freezing dynamic clips and animating static ones, so we also report the mean per-clip gap
\begin{equation}
  \Delta\mathrm{St} = \frac{1}{|\mathcal{V}|}\sum_{v\in\mathcal{V}}
  \bigl|\widehat{\mathrm{St}}(v) - \mathrm{St}(v)\bigr|,
\end{equation}
where $\mathcal{V}$ is the set of clips for which both scores are defined. 

\begin{table}[t]
\centering
\caption{\textbf{\worldupdate{} consistency across different model choices.} Models are conditioned with ground truth neighbor views reprojections. SCS denotes the Scene Consistency Score, measuring static (SCS-st) and dynamic (SCS-dyn) scene content consistency.}
\label{tab:world_update_consistency_2}
\setlength{\tabcolsep}{3.2pt}
\begin{tabular}{l*{6}{c}}
\multicolumn{1}{c}{\bf Method}
& \bf PSNR $\uparrow$
& \bf LPIPS $\downarrow$
& \bf SCS-static $\uparrow$
& \bf SCS-dyn $\uparrow$
& \bf Stat $\downarrow$
& \bf Stat Gap $\downarrow$ \\
\midrule
\multicolumn{7}{l}{\textit{test 1: seen towns}} \\
Persistent Baseline
& \textbf{33.66}
& \textbf{0.118}
& 0.807
& \textbf{0.208}
& 0.993
& 0.565 \\
Dynamically Grounded WM
& 30.96
& 0.166
& 0.800
& 0.193
& 0.870
& 0.480 \\
Multi-Rig DGWM
& 30.06
& 0.191
& \textbf{0.841}
& 0.124
& \textbf{0.498}
& \textbf{0.204} \\
\midrule
\multicolumn{7}{l}{\textit{test 2: unseen town}} \\
Persistent Baseline
& \textbf{31.88}
& \textbf{0.110}
& \textbf{0.801}
& \textbf{0.228}
& 0.985
& 0.593 \\
Dynamically Grounded WM
& 27.44
& 0.184
& 0.723
& 0.186
& 0.914
& 0.519 \\
Multi-Rig DGWM
& 25.99
& 0.214
& 0.730
& 0.097
& \textbf{0.427}
& \textbf{0.234} \\
\end{tabular}%
\end{table}
\section{Additional Experiments}
\subsection{\worldupdate{} Consistency}
In this experiment we evaluate the multi-view abilities of our WorldUpdate model for generating video across rigs and views. To this end, we take the two closest to the current camera orientations of the neighboring rigs and reproject them to the current view. We evaluate how close the model predictions are to the ground truth of those views at that time. We compare our multi-view WorldUpdate with WorldInteract, which is dynamically grounded but not multi-view in design.

Results are shown in Tab.~\ref{tab:world_update_consistency_2}. A baseline that freezes the conditioning frame achieves the best PSNR and LPIPS. This is a property of the setting: the rig cameras are stationary, so most of the image is static in time and visual metrics reward generating no motion. SCS-dyn is affected in the same way. It is an overlap score, and over one evaluation step a moving object still overlaps most of itself, so a frozen frame is credited with the overlap the object retains, while a model that moves it is penalised wherever its motion departs from the single recorded future. Both must therefore be read together with the staticness gap. \worldinteract{} stays close to the frozen baseline in staticness, and inherits its PSNR, LPIPS and SCS-dyn from that. \worldupdate{} has the lowest staticness gap in every configuration, moving the scene by roughly the amount the ground truth does, and pays for it on every metric that rewards freezing, while preserving static structure best of all generative models (SCS-st) and exceeding the frozen baseline on it in the seen towns. This experiment thus exposes a trade-off between a preference for static content and dynamic accuracy across models, and a bias of the metrics themselves towards the former.

\end{document}